\documentclass[10pt,twocolumn,letterpaper]{article}

\usepackage[pagenumbers]{cvpr} 

\usepackage[dvipsnames]{xcolor}
\usepackage{colortbl}
\usepackage{array}
\usepackage{color}
\usepackage{multirow}
\usepackage{amsfonts}
\usepackage{bbm}
\usepackage{graphicx}
\usepackage{booktabs}
\usepackage{dsfont}
\definecolor{darkgreen}{RGB}{0, 150, 0} 
\definecolor{lightgray}{gray}{0.9}
\definecolor{mediumgray}{gray}{0.7}
\usepackage{amssymb}
\usepackage{pifont}
\usepackage{bm}
\usepackage{wrapfig}
\usepackage{overpic}

\newcommand{\methodName}{\emph{ContextSeg}\xspace}

\definecolor{cvprblue}{rgb}{0.21,0.49,0.74}
\usepackage[pagebackref,breaklinks,colorlinks,citecolor=cvprblue]{hyperref}

\newcommand{\parag}[1]{{\vspace{1mm} \noindent \textbf{#1}}}
\newcommand{\rev}[1]{{\color{black}#1}}

\title{\methodName: Sketch Semantic Segmentation by Querying the Context \\ with Attention}

\author{Jiawei Wang\\
Shandong University
\and
Changjian Li\\
The University of Edinburgh
\and
\href{https://enigma-li.github.io/projects/contextSeg/contextSeg.html}{https://enigma-li.github.io/projects/contextSeg/contextSeg.html}
}

\begin{document}

\maketitle

\begin{abstract}
Sketch semantic segmentation is a well-explored and pivotal problem in computer vision involving the assignment of pre-defined part labels to individual strokes. 
This paper presents \methodName -- a simple yet highly effective approach to tackling this problem with two stages.
In the first stage, to better encode the shape and positional information of strokes, we propose to predict an extra dense distance field in an autoencoder network to reinforce structural information learning. 
In the second stage, we treat an entire stroke as a single entity and label a group of strokes within the same semantic part using an auto-regressive Transformer with the default attention mechanism. 
By group-based labeling, our method can fully leverage the context information when making decisions for the remaining groups of strokes.
Our method achieves the best segmentation accuracy compared with state-of-the-art approaches on two representative datasets and has been extensively evaluated demonstrating its superior performance.
Additionally, we offer insights into solving part imbalance in training data and the preliminary experiment on cross-category training, which can inspire future research in this field.
\end{abstract}
    
\section{Introduction}
\label{sec:intro}

\begin{figure}[!t]
    \centering
    \includegraphics[width=1.0\linewidth]{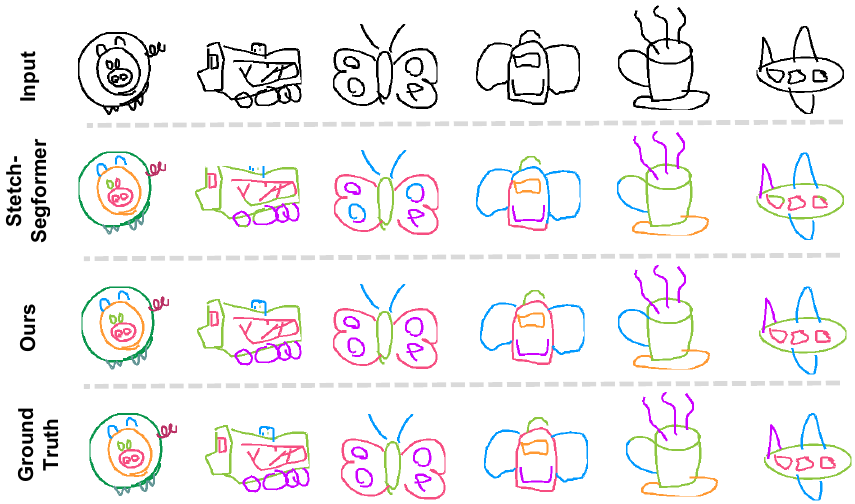}
    \vspace{-6mm}
    \caption{Given an input sketch, semantic segmentation is to assign labels to strokes based on their semantics so as to form semantic groups. Our method is robust to stroke variations achieving superior results (e.g., the correctly labeled airplane windows).
    }
    \vspace{-5mm}
    \label{pic:teaser}
\end{figure}

Sketches are widely used as a human-computer interaction tool. Many studies keep exploring its capabilities in modeling~\cite{li2022free2cad, li2018robust}, retrieval~\cite{bui2018sketching,bui2017compact} and generation~\cite{cheng2023adaptively,bhunia2022doodleformer}. Among these tasks, sketch interpretation serves as the foundation. However, it still remains challenging for computers due to the inherent ambiguity and sparsity of user sketches~\cite{yang2021sketchgnn}. In this paper, we focus on sketch semantic segmentation, an essential task in finer-level sketch interpretation.

Sketches are typically represented in three data formats, i.e., the raster image, graph and point sequence.
Recently, many learning-based approaches have adopted the aforementioned sketch representations revealing various advantages and disadvantages (\rev{see the inset table}).
Specifically, image-based methods~\cite{li2018fast,zhu2018part,zhu20202d} take raster images as input, exploiting absolute coordinates to  capture the
\begin{wrapfigure}[5]{r}{0.6\linewidth}
\vspace{-3mm}
    \begin{overpic}[width=1.0\linewidth]{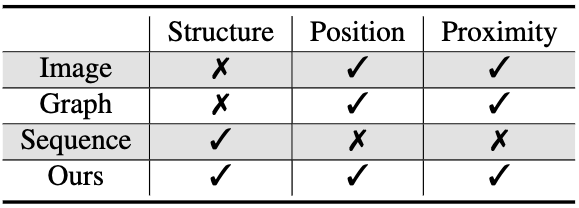}
    \end{overpic}
\end{wrapfigure}
proximity between strokes. However, they tend to perform poorly in learning the structural information about strokes and overlook the sequential relationships between them. 
Graph-based methods~\cite{yang2021sketchgnn,wang2020multi,zheng2023sketch} take graphs as input, which consists of stroke points as nodes and point connections as edges. The graph representation also exploits absolute coordinates, which facilitate proximity learning between strokes but is challenging to capture the structural information of a stroke. 
Sequence-based methods~\cite{ha2017neural,wu2018sketchsegnet,qi2019sketchsegnet+,li2019toward} take as input point sequences, which use relative coordinates to better encode structural information of individual strokes. However, they struggle to capture the proximity relationships and spatial information between strokes. 
Nevertheless, stroke structural and positional information and the proximity relationship between strokes are essential for sketch interpretation.

Motivated by the observations, we propose \methodName, a two-stage approach marrying advantages of all three representations \rev{to capture better structural and positional information of each single stroke, and the proximity relationship between strokes}. Specifically, at the stroke level, we treat an entire stroke instead of the stroke point as a single entity, represented as a raster image, where the positional information can be efficiently extracted by an autoencoder CNN network. Furthermore, due to the sparsity of stroke pixels, as well as the large geometric variation of a stroke depicting the drawing content, it is challenging to encode stroke structural information only by reconstruction. We thus propose a novel dense distance field regression task upon reconstruction, reinforcing structural information learning.
At the sketch level, we treat all strokes within a sketch as a sequence of images. By augmenting the sequence with positional information, neural networks can well capture the proximity between strokes.
As for the segmentation task, we exploit an auto-regressive Transformer, where instead of labeling one stroke at each time, we select a set of strokes belonging to the same semantic group at one time. Using the group-based prediction, context information, e.g., strokes possessing a part label and strokes remaining to be labeled, can be explicitly utilized by the Transformer decoder when making decisions for remaining strokes.

To demonstrate the effectiveness of our network, we have conducted experiments on two representative datasets, i.e., SPG~\cite{li2019toward} and CreativeSketch~\cite{ge2020creative}. Experimental results including comparisons and ablation studies demonstrate our superior performance.

In summary, our main contributions are threefold:
\begin{itemize}
    \item We propose a CNN-based network for stroke embedding learning, which features dense distance field prediction for capturing the structural information of a stroke. 
    \item We exploit an auto-regressive Transformer network for segmentation, where we propose to label a group of strokes at one time fully leveraging context information leading to our SOTA performance.
    \item We propose a novel semantic-aware data augmentation mechanism attempting to address the data imbalance problem; and in benefitting from our network's capability to extract and exploit contextual information, we achieved significant performance improvement in the challenging cross-category learning. Both the novel strategy and the preliminary experiment might inspire future research.
\end{itemize}

\section{Related Work}
\label{sec:related_work}
\parag{Sketch Representation.} In the field of computer vision, sketch representation learning has garnered widespread attention. \rev{It is a fundamental task in numerous downstream applications, i.e.,  sketch-based image retrieval (SBIR) \cite{sain2021stylemeup, bui2018sketching}, sketch generation \cite{ribeiro2020sketchformer}, and sketch classification \cite{zhang2016sketchnet}.} Zhang et al. \cite{zhang2016sketchnet} introduced deep convolutional neural networks for learning sketch embedding. The network took triplets composed of sketches, positive real images, and negative real images as input to discover coherent visual structures between the sketch and its positive pairs. \rev{Sain et al. \cite{sain2021stylemeup} employed a cross-modal VAE to disentangle sketches into shared semantic content and unique style contents, enhanced with meta-training for dynamic adaptation to unseen user styles, enabling style-agnostic SBIR.} Furthermore, research has been conducted on the generalization properties of sketch embedding \cite{pang2019generalising, liu2019sketchgan, bhunia2022doodleformer, ge2020creative}. 

\parag{Semantic Sketch Segmentation.} 
Compared to previous works that relied on handcrafted features and complex models \cite{sun2012free, schneider2016example,delaye2015flexible}, many deep learning-based models have achieved outstanding performance in sketch semantic segmentation tasks \cite{yang2021sketchgnn,wu2018sketchsegnet,zheng2023sketch}. Based on the data formats they use, existing methods can be broadly grouped into three classes: image-based \cite{li2018fast,zhu2018part,zhu20202d}, sequence-based \cite{wu2018sketchsegnet,qi2019sketchsegnet+, li2019toward}, and graph-based \cite{yang2021sketchgnn,wang2020multi,zheng2023sketch} methods. \rev{Zhu et al. \cite{zhu2018part} proposed a dual-CNN approach for sketch segmentation and labeling, employing two networks with distinct kernel sizes to handle different sketch lengths, while enhancing performance through the integration of position and orientation as a triple-channel input using fused masks.} Li et al. \cite{li2019toward} proposed a sequence-based encoder-decoder architecture for sketch semantic segmentation, applying additional constraints in the loss function, specifically targeting reconstruction and global grouping consistency. Yang et al. \cite{yang2021sketchgnn} proposed a graph-based network and utilized two graph convolutional branches to extract the inter-stroke features and the intra-stroke features. 

\begin{figure*}[!h]
    \centering
    \includegraphics[width=1\textwidth]{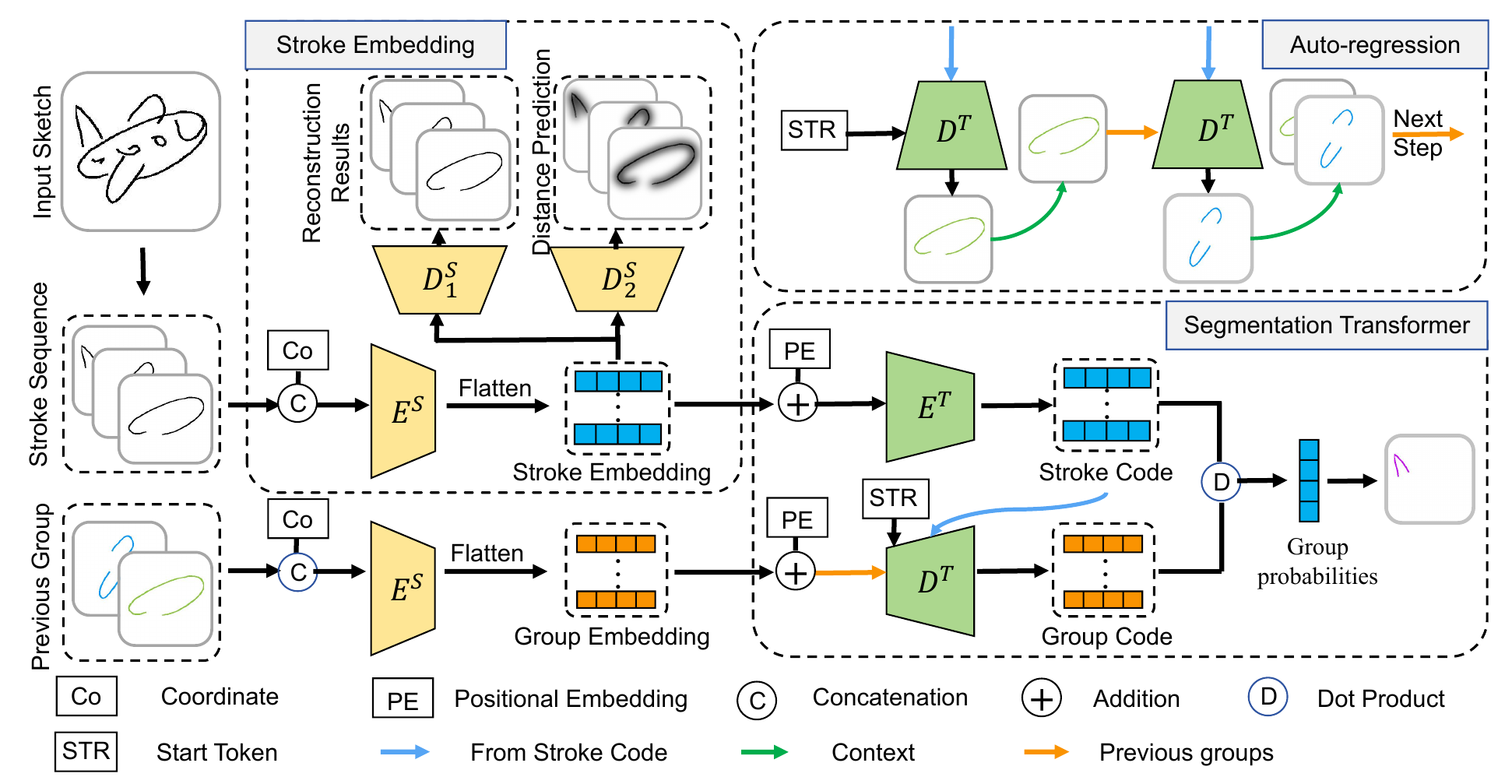}
    \vspace{-7mm}
    \caption{Overview of \methodName. Given an input sketch, it is first divided into a sequence of strokes, which are used to train our stroke embedding network -- an autoencoder with an extra distance field output (Sec.~\ref{subsec:stroke_embedding}). Then, the learned embeddings are sent to the segmentation Transformer operating in an auto-regressive manner (Sec.~\ref{subsec:seg_trans}). The Transformer leverages contextual information, encompassing previously labeled strokes and remaining strokes, as input for the current step's stroke labeling. 
    }
    \vspace{-4mm}
    \label{fig:pipeline}
\end{figure*}

\parag{Vision Transformer.}
\rev{Other than natural language processing, Transformer \cite{vaswani2017attention} has been widely adopted in all kinds of computer vision tasks, e.g., classification \cite{dosovitskiy2020image,fan2021multiscale}, detection \cite{carion2020end}, segmentation \cite{liu2021swin}.
}
Unlike the auto-regressive decoding strategy, Carion et al. \cite{carion2020end} proposed parallel decoding, given the absence of inherent order or sequence information between different bounding boxes in object detection tasks. Transformer was also adopted in the sketch domain, where Ribeiro et al. \cite{ribeiro2020sketchformer} encoded free-hand sketches in a vector format and effectively improved the performance of sketch-based image retrieval while Li et al. \cite{li2022free2cad} treated sketch-based CAD modeling as a serializable translation problem and employed a Transformer-based network for stroke grouping. 
\section{Method}
\label{sec:method}
Given an input sketch $\mathbf{S}$ represented by a sequence of strokes $\{s_i\}$, our goal is to segment them into groups based on their underlying semantics and assign the corresponding semantic label to the group of strokes. Figure~\ref{fig:pipeline} displays an overview of our method, which consists of two key modules, i.e., \emph{stroke embedding} and \emph{segmentation Transformer}. In what follows, we elaborate on the details.

\subsection{Stroke Embedding}
\label{subsec:stroke_embedding}
We employ a Transformer as our segmentation model, which accepts embedding vectors. To this end, given a stroke $s_i$, we first design our embedding learning network to obtain the corresponding stroke embedding $s^e_i$. 
Specifically, we represent each stroke $s_i$ as a binary image $I(s_i) \in\{0,1\}^{256 \times 256}$. 
Since we want the learned embedding to encode stroke positional information, we thus augment the input image channel with two additional coordinate channels following the practice in CoordConv~\cite{liu2018intriguing}.
Given the coordinate augmented stroke image, we leverage a 2DCNN-based autoencoder network as our backbone to extract stroke embedding. Both the encoder $E^S$ and decoder $D^S_1$ consist of a few 2DCNN layers with varying convolution kernels and feature dimensions. After the last layer of the encoder, we flatten the feature to obtain the embedding $s^e_i$. Note that, skip-connections are omitted for embedding learning.
Ideally, through minimizing the disparity between the input $I(s_i)$ and the reconstructed image $D^S_1(E^S(I(s_i))$, the trained network excels in reconstructing the input accurately and encoding the stroke effectively with high fidelity. 

\begin{figure*}[!t]
	\centering
    \includegraphics[width=1\textwidth]{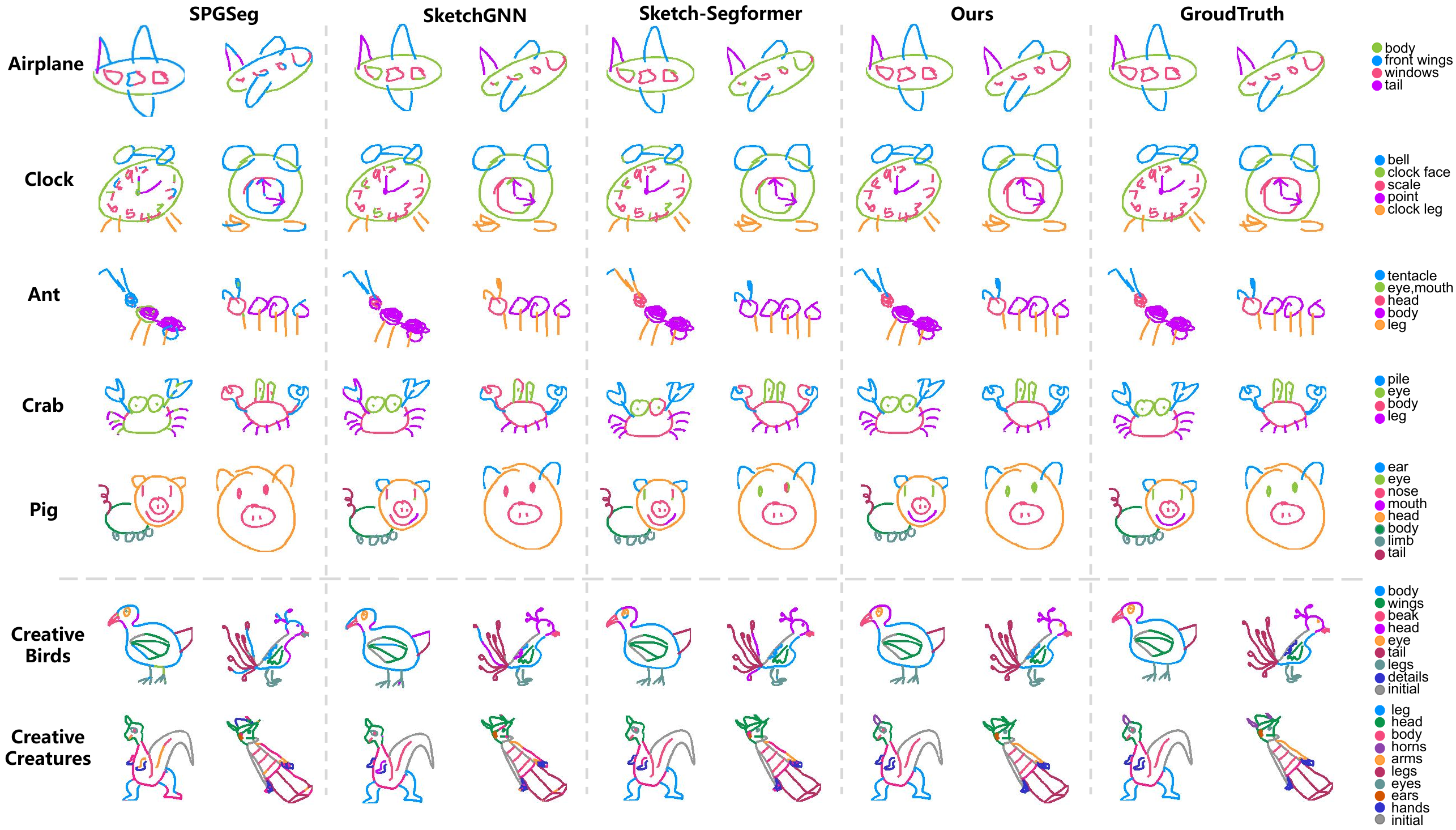}
    \vspace{-7mm}
	\caption{Visual comparison with three competitors on the SPG and the CreativeSketch datasets.}
    \vspace{-4mm}
	\label{pic:main_result2}
\end{figure*}

However, unlike photorealistic images, a sketch is usually harder to encode due to its sparsity and large variation of the compositional strokes. For example, some strokes might be cluttered in a small region due to over-sketching (see the body of an ant in Fig.~\ref{pic:main_result2}) and an individual stroke might exhibit large geometric variations based on the drawn content (see the wing of the angel in Fig.~\ref{pic:ablation}). These challenges fail the aforementioned baseline embedding network (Sec.~\ref{subsec:abl_study}, ablation study). Consequently, to address these difficulties, we design a novel distance field regression in addition to the reconstruction in the autoencoder.

\begin{figure}[!tb]
    \centering
    \includegraphics[width=0.85\linewidth]{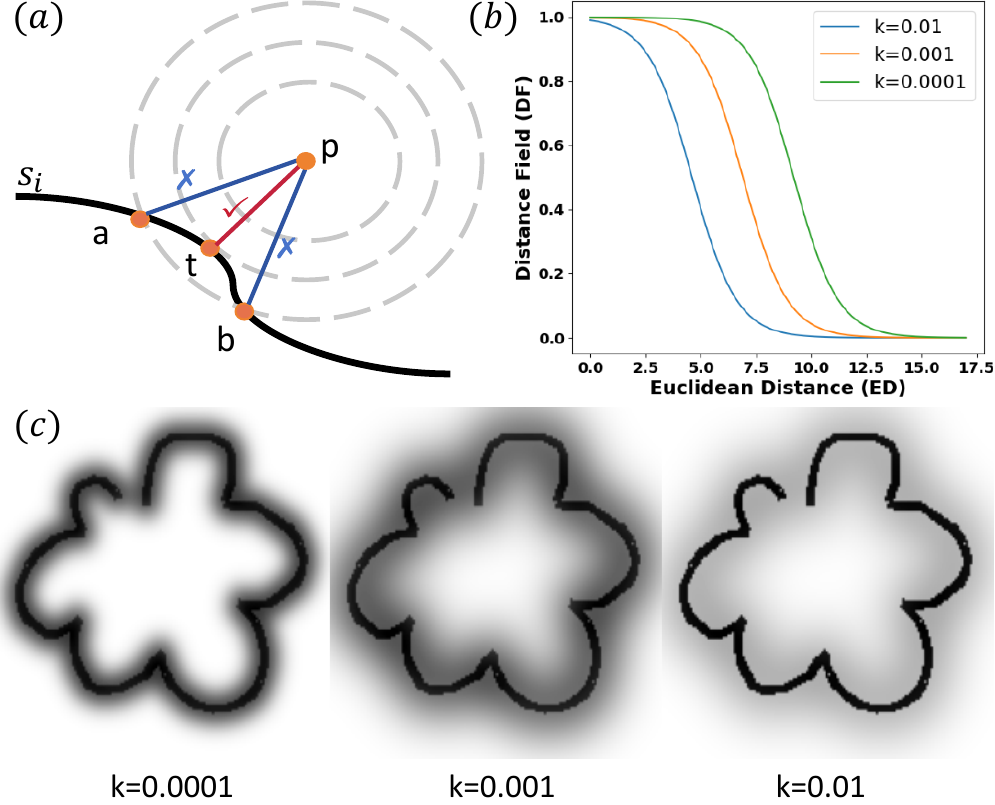}
    \vspace{-2mm}
    \caption{Stroke distance field. (a) Given an arbitrary point $p$ in the image, we calculate the shortest Euclidean distance from $p$ to the point $t$ on the stroke. (b) The distance curves with three different $k$ values. (c) Distance field maps of three typical $k$ values.
    }
    \vspace{-4mm}
    \label{pic:dis_func}
\end{figure}

\parag{Stroke Distance Field.} Since we treat each stroke as an image $I(s_i)$, a natural choice to build the distance field is the unsigned distance from arbitrary points in the image to the stroke. However, with a large region expanding the whole image into consideration, the structural information of the stroke is overlooked. 
We thus propose a novel distance field (dubbed as $DF^s$) concentrating only a narrow band around the stroke, defined over all the points on the image:
\begin{equation}
    DF^s(s_i) = \frac{1}{1+k \times e^{dis^{p}_{s_i}}}
\end{equation}
where $dis^{p}_{s_i}$ calculates the shortest Euclidean distance from a query point $p$ on the image to the stroke $s_i$ (Fig.~\ref{pic:dis_func}(a)), while $k$ controls both the width of the narrow band as well as the decay rate of distance values after leaving the stroke (Fig.~\ref{pic:dis_func}(b)). 
As shown in Fig.~\ref{pic:dis_func}(c), we experimentally set $k$ at $0.001$, producing a dense field within an appropriate band and the surrounding values depicting the stroke variation. 

The advantages of using $DF^s$ are two-fold. On the one hand, by only attending the narrow band around the stroke with a smooth value decay, the structural information of the stroke is forced to be captured. On the other hand, the distance function maps any Euclidean distance to $[0,1]$, which is easy to be regressed by neural networks.

\parag{Network Configuration and Training.} We thus augment the baseline autoencoder with a second decoding branch (i.e., $D^S_2$ in Fig.~\ref{fig:pipeline}) to output a dense distance field respecting the input stroke. To train the embedding network, we minimize the following loss function:
\begin{equation}
    \mathcal{L}_{em} = \mathcal{L}_{recon} + \gamma\mathcal{L}_{dis},
\end{equation}
where $\gamma$ is a balancing weight, is set at $0.5$ empirically. $\mathcal{L}_{recons}$ is the image reconstruction loss, defined as:
\begin{equation}
    \mathcal{L}_{recons} =\left\|I\left(s_{i}\right)-D_{1}^{S}\left(E^{S}\left(I\left(s_{i}\right)\right)\right)\right\|^{2},
\end{equation}
and $\mathcal{L}_{dis}$ is defined as:
\begin{equation}
    \mathcal{L}_{dis} = \left\|DF^s\left(s_{i}\right)-D_{2}^{S}\left(E^{S}\left(I\left(s_{i}\right)\right)\right)\right\|^{2}.
\end{equation}
For any stroke image $I(s_i)$, we first calculate the distance map $DF^s(s_i)$ as a pre-processing step (see three examples in Fig.~\ref{pic:dis_func} (c)). After training the network, we simply discard the two decoders and fix the encoder to obtain the stroke embedding $s_i^e = E^S(I(s_i))$.

\parag{Embedding of a Group of Strokes.} Other than the single strokes, we also encode a set of strokes corresponding to a semantic group, which is used in our segmentation Transformer. Specifically, suppose $g_i$ is a group of strokes, and $I(g_i)$ is the corresponding image. We use all these group images as extra training data for the aforementioned embedding network training and obtain the group embedding \rev{$g_i^e=E^s(I(g_i))$} as same as for a single stroke. 

\subsection{Segmentation Transformer}
\label{subsec:seg_trans}

As shown in Fig.~\ref{fig:pipeline}, our segmentation Transformer is built upon~\cite{vaswani2017attention}, running in an auto-regressive manner with self- and cross-attention mechanisms in the encoder $E^T$ and $D^T$. Specifically, given the set of stroke embeddings $\{s_i^e\}$, we first augment them with the sinusoidal positional embedding to help distinguish one from the others. Then, the encoder $E^T$ takes as input a sequence of augmented stroke embeddings and applies a few self-attention layers to produce the set of stroke codes $\{\mathbf{s_i}\}$, which servers as one of the inputs to the decoder $D^T$.

As for the decoding, we use the auto-regressive scheme, i.e., when making the decision of the current iteration step, the context information from the previous predictions serves as input to the decoder. The auto-regression will stop until nothing remains to be decided for the current step.
In our case, the context information is the labeled strokes in the previous steps and the strokes remain to be labeled. 
In each decoding iteration, a possible solution is to assign a label to the current stroke. However, even with context information, this stroke-based decoding ignores the group information. Instead, inspired by the Pointer Networks~\cite{vinyals2015pointer}, our decoder outputs a group code $\mathbf{g_j}$ each time by considering the labeled groups of strokes as context, and the group code is leveraged as a pointer to select strokes belonging to the current semantic group by dot product comparison leading to a group probability. The selection process is defined as:
\begin{equation}
p_i^j = \sigma\left(\mathbf{s_i} \cdot \mathbf{g_j}\right),
\end{equation}
where $( \cdot )$ is the dot produce, and $\sigma$ is the sigmoid activation. In practice, when the probability $p_i^j$ is greater than 0.5, we consider the stroke $s_i$ belonging to the $j$-th group.

\parag{Network Training.} To train the network, we can simply minimize the cross-entropy loss \cite{zhang2018generalized} applied to a group of strokes at one time. However, we observe that the training data has a severe imbalance in terms of the stroke distribution over all semantic groups, e.g., all the ants have the body and leg groups, each with many strokes, while only a few of them have the eye group with one or two strokes. We thus employ the focal loss to train our segmentation Transformer, defined as:
\begin{equation}
    \begin{split}
        \mathcal{L}_{seg} = \sum_{g_{j}} \sum_{s_{i}} -\mathds{1}_{s_{i} \in g_{j}}\left( 1-p_{i}^{j}\right)^{\gamma}\log \left(p_{i}^{j}\right)\\
        -\left(1-\mathds{1}_{s_{i} \in g_{j}}\right)\left( {p_{i}^{j}}\right)^\gamma \log \left(1-p_{i}^{j}\right),
    \end{split}
\end{equation}
where $\mathds{1}$ is the indicator function, and $\gamma$ is the focusing parameter ($2.0$ in our case) in the modulating factor.

\section{Experiments}
\label{sec:exps}

In what follows, we present visual and statistical evaluations and provide implementation details in the supplementary.

\parag{Datasets.} Following existing work, we use both the SPG~\cite{li2019toward} and CreativeSketch~\cite{ge2020creative} (dubbed as CS) datasets in this paper. 
Specifically, the SPG dataset contains sketches from 25 categories, each with 800 sketches. As in~\cite{li2019toward,yang2021sketchgnn}, we use the same 20 selected categories and employ the same data splitting as described in \cite{wu2018sketchsegnet}, i.e., 700 sketches for training, 100 sketches for testing. We use all the training data from 20 categories for the embedding network training, while the segmentation Transformer is trained per category. As described in Sec.~\ref{subsec:stroke_embedding}, we collected both single strokes and groups of strokes to train the embedding network.
As for the CS dataset, it consists of 2 categories with much more complex sketches forming 8 and 17 semantic parts, respectively. We then randomly select 3000 sketches from each category with a 2500-500 training and testing split. Similarly, we use all 5000 sketches from both categories to train the embedding networks, while training the segmentation Transformer per category. All the sketches are scaled to fit into $256\times256$ images. 

\parag{Evaluation Metrics.}
We have adopted three metrics~\cite{huang2014data,wu2018sketchsegnet,li2022free2cad} to evaluate performance, i.e., stroke accuracy (SAcc), grouping accuracy (GAcc) and component accuracy (CAcc), and their definitions can be found in the supplementary. Intuitively, stroke accuracy provides a fine-grained evaluation inspecting each stroke, while both the grouping accuracy and component accuracy quantify the semantic-based accuracy of all strokes. Compared with GAcc, CAcc is more vulnerable to the long tail problem regarding the stroke distribution over semantic groups.

\begin{table*}[!t]
\caption{Statistical comparison with SOTA methods on the SPG and CreativeSketch datasets, all three metrics are reported.}

\vspace{-2mm}

\fontsize{5}{4}\selectfont
\label{tab:main_stats}
\renewcommand{\arraystretch}{1.6}
\resizebox{1\textwidth}{!}{
\begin{tabular}{c| c | c c c |c c c | c c c | c c c }

\toprule[0.3mm]

&\multirow{2}*{ \textbf{Category}} & \multicolumn{3}{c|}{\textbf{SPGSeg}\cite{li2019toward}} & \multicolumn{3}{c|}{\textbf{SketchGNN}\cite{yang2021sketchgnn}} & \multicolumn{3}{c|}{\textbf{Sketch-Segformer}\cite{zheng2023sketch}} &\multicolumn{3}{c}{\textbf{Ours}} \\
\cline{3-14}

& & \textbf{SAcc} & \textbf{GAcc} & \textbf{CAcc}  & \textbf{SAcc} & \textbf{GAc}c & \textbf{CAcc}  & \textbf{SAcc} & \textbf{GAcc} & \textbf{CAcc}
& \textbf{SAcc} & \textbf{GAcc} & \textbf{CAcc}\\

\hline

\multirow{21}*{ \rotatebox{90}{\textbf{SPG}}} & Airplane &72.1 &  85.9 & 70.1 & 91.1 & 94.3 & 86.2 & 92.4 &  94.4 & 87.4 &  \textbf{93.2}  & \textbf{94.9} & \textbf{89.5}\\
& Alarm clock &82.4
 &  89.5 & 73.5 & 95.9 & 97.1 & 91.0 &95.7&  96.4 & 90.6  & \textbf{96.3} & \textbf{97.6} & \textbf{93.3} \\

& Ambulance &69.3 &  84.6 & 61.6 & 90.6 & 94.8 & 81.9 &91.4 &  95.1 & 82.3  & \textbf{92.2} & \textbf{95.6} & \textbf{86.0} \\
& Ant&58.9 &  85.4 & 51.5 & 91.4 & 95.4 & 82.1 &92.7 & 95.9 & 82.4 & \textbf{94.8} & \textbf{96.2} & \textbf{86.1} \\

& Apple &75.9 &  83.9 & 69.5 & 91.1 & 94.1 & 89.5 &92.5&  94.6 & 90.1  & \textbf{94.2} & \textbf{94.9} & \textbf{92.7} \\
& Backpack &64.8 &  84.6 & 59.5 & 84.1 & 93.5 & 79.3 &85.3 &  93.7 & 80.4  & \textbf{87.8} & \textbf{94.5} & \textbf{83.9} \\

& Basket &83.1 &  86.5 & 73.9 & 95.9 & 97.4 & 93.9 &96.7 &  96.8  & 93.5 & \textbf{97.8} & \textbf{98.0} & \textbf{94.2} \\
& Butterfly &82.4 &  87.5 & 79.5 & 97.9 & 98.1 & 96.1  &97.7 &  97.9 &95.9 & \textbf{98.8} & \textbf{98.6} & \textbf{96.5} \\

& Cactus &76.7 &  89.9 & 75.4 & 95.2 & 97.3 & 93.3  & 96.1 &  97.4  &  93.4 & \textbf{96.9} & \textbf{97.7} & \textbf{93.8} \\
& Calculator &88.7 &  90.5 & 77.4 & 98.4 & 98.3 & 97.2  &98.5 & 98.3 & 97.1  & \textbf{99.2} & \textbf{98.7} & \textbf{97.5} \\

& Campfire &90.2 &  93.2 & 79.5 & 95.3 & 96.6 & 92.5  & 96.7  & 96.8 & 92.7& \textbf{97.1} & \textbf{97.1} & \textbf{92.8} \\
& Candle &74.9 &  87.6 & 65.9 & 97.6 & 98.1 & 95.8 &98.1 &   98.3 & 95.9 &\textbf{ 98.7 }& \textbf{98.5 }& \textbf{96.0}\\

& Coffee cup &86.6 &  88.9 & 83.6 & 97.8 & 98.4 & 94.5  &98.3 & 98.6 &\textbf{94.7} &\textbf{ 98.9} & \textbf{98.8} & \textbf{94.7 }\\
& Crab &71.3 &  84.9 & 69.3 & 94.0 & 95.8 & 87.9  &93.9 &   96.4 & 89.9&\textbf{ 95.6} &\textbf{ 96.4} & \textbf{91.4} \\

& Duck &74.2 &  83.5 & 68.5& 96.3& 97.3 & 90.6  &96.1 &  96.1& 90.2 & \textbf{97.3} & \textbf{97.8} & \textbf{93.0} \\
& Face &81.9&  87.5 & 83.6 & 96.2 & 98.2 & 94.3 &\textbf{97.2} & \textbf{ 98.6 }& \textbf{94.9} & 96.4 & 98.4 & 94.8 \\

& Ice cream &80.5 &  86.9 & 88.4 & 94.5 & 95.6 & 91.0 &94.4 &  95.6& 90.9 & \textbf{95.9} & \textbf{96.5} & \textbf{92.4} \\
& Pig &77.9 &  83.7 & 74.9 & 96.3 & 98.1 & 94.9 &96.1 & 98.2 & 94.8 & \textbf{97.8} & \textbf{98.8} & \textbf{95.6 }\\

& Pineapple &91.5 &  92.7 & 85.6 & 95.2 & 96.2 & 91.0 &96.3 & 96.7 & 91.6 & \textbf{97.3} & \textbf{97.0 }& \textbf{92.5 }\\
& Suitcase &91.9 &  93.1 & 87.6 & 96.4 & 97.8 & 95.1 &97.3 & 98.3 & 95.6 & \textbf{97.6} & \textbf{98.6} & \textbf{95.9 }\\
\cline{2-14}

& Average &78.8 &  87.5 & 73.9 & 94.6 & 96.6 & 90.9 &95.2 &  96.7 & 91.2  & \textbf{96.2} & \textbf{97.2 }& \textbf{92.6} \\
\hline
\hline

\multirow{3}*{ \rotatebox{90}{\textbf{CS}}} & Birds &56.4 &81.5  & 37.2&  67.4 & 85.3 & 43.4 & 68.7  &86.5   & 45.0 & \textbf{70.2}  &\textbf{89.3} & \textbf{65.8}\\
& Creatures &37.8 &76.5  & 28.5 &46.6 &80.4  & 41.6 & 47.2& 81.3 & 42.7 &\textbf{50.1} & \textbf{85.6} &\textbf{45.8}  \\
\cline{2-14}

& Average & 47.1&79.0   & 32.9 & 57.0 & 82.9 &42.5 & 58.0& 83.9 & 43.9  &\textbf{60.2}  & \textbf{87.5}&  \textbf{55.8}\\
\bottomrule[0.3mm]
\end{tabular}
}
\vspace{-2mm}
\end{table*}


\begin{table*}[!htb]
\centering
\caption{Quantitative results of our ablation study on the SPG dataset.}
\vspace{-2mm}
\label{tab:ablation}
\renewcommand{\arraystretch}{1.3}
\resizebox{1\textwidth }{!}{
\begin{tabular}{c|ccc|ccc|ccc|ccc|ccc}
\toprule[0.4mm]
\multirow{2}{*}{Method} & \multicolumn{3}{c|}{Airplane}& \multicolumn{3}{c|}{Calculator}& \multicolumn{3}{c|}{Face }& \multicolumn{3}{c|}{Ice cream}& \multicolumn{3}{c}{Average} \\
\cline{2-16}

& SAcc& GAcc & CAcc& SAcc& GAcc & CAcc& SAcc& GAcc & CAcc& SAcc& GAcc & CAcc & SAcc& GAcc & CAcc\\

\hline

\rowcolor{lightgray}
Sketchformer-Seg&  76.4 & 84.6 & 64.7 & 82.5 & 87.5 & 78.4& 75.4 & 84.2& 69.1& 77.3 & 85.4 & 62.1 & 77.9 & 85.4 & 68.6\\
Ours w/o CC & 85.6  &87.5 & 82.4 & 91.1& 92.1 & 89.2 & 91.7 & 90.3  & 87.3& 87.9 & 90.1 & 86.4 & 89.1 & 90.0 & 86.3\\

\rowcolor{lightgray}
Ours w/o DF & 90.9  &92.4 & 86.5 & 96.7 & 97.1 & 94.5 & 94.6 & 95.1& 92.3& 93.4 &94.5&90.4 &93.9 &94.8 & 90.9\\
Ours-PD & 91.1 & 92.5 & 86.8 &97.2 &97.8 &95.1 &95.3 & 96.2& 93.1 & 94.1 & 95.3 & 91.2& 94.4 & 95.5 & 91.6 \\
\hline

\rowcolor{lightgray}
Ours & \textbf{93.2} &\textbf{94.9} & \textbf{89.5}& \textbf{99.2} &\textbf{98.7} & \textbf{97.5}& \textbf{96.4}& \textbf{98.4} & \textbf{94.8} &\textbf{95.9}& \textbf{96.5} & \textbf{92.4} &\textbf{96.2} & \textbf{97.1}
& \textbf{93.6} \\
\bottomrule[0.4mm]
\end{tabular}
}
\end{table*}

\subsection{Comparison}
\label{subsec:comp}
We compare our approach with three SOTA methods, i.e., SPGSeg~\cite{li2019toward}, SketchGNN~\cite{yang2021sketchgnn} and Sketch-Segformer~\cite{zheng2023sketch}. All three methods are trained from scratch using their default parameters on our dataset. In the following, we introduce the quantitative and qualitative results.

Quantitative comparisons are shown in Tab.~\ref{tab:main_stats}, wherein we obtain the best average and per-category results (except for the face category) over all three metrics on both the SPG and CS datasets. Specifically, in terms of component accuracy, our approach achieved notable advancements of $1.4\%$, $1.7\%$, and $18.7\%$ compared to Sketch-Segformer, SketchGNN, and SPGSeg, respectively, on the SPG dataset. Also, on the more challenging CS dataset, substantial improvements in component accuracy were observed, reaching $11.9\%$, $13.3\%$, and $22.9\%$ in comparison to Sketch-Segformer, SketchGNN, and SPGSeg, respectively. 
Regarding the face category, our method demonstrates comparable accuracies, albeit slightly lower. This outcome is attributed to the presence of highly overlapped strokes, particularly representing hairs and eyes, which pose challenges to precise embedding learning and subsequently impact grouping predictions.
In Fig.~\ref{pic:main_result2}, qualitative results are showcased, illustrating our method's capability to successfully identify strokes delineating intricate semantic details. Notably, it accurately captures elements such as clock numbers, as well as the wings and tail of a bird. More analysis and failure cases can be found in the supplementary.

\begin{figure*}[!htb]
    \centering
    \includegraphics[width=0.95\textwidth]{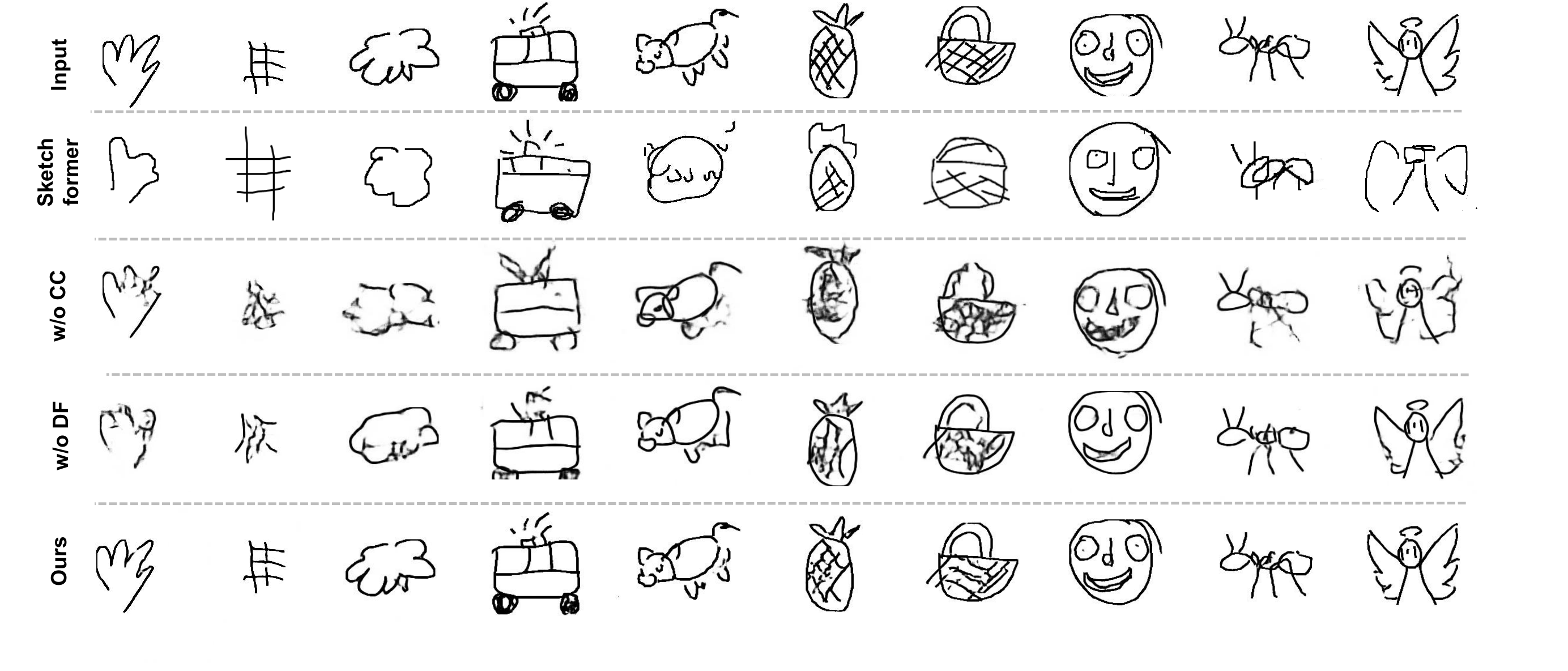}
    \vspace{-2mm}
    \caption{Sketch reconstruction results of our ablation study on different stroke embedding networks. 
    }
    \label{pic:ablation}
    \vspace{-4mm}
\end{figure*}

\subsection{Ablation Study}
\label{subsec:abl_study}

\parag{Stroke Embedding.} We modify the stroke embedding module, leading to three alternative embedding networks:
\begin{itemize}
    \item Sketchformer~\cite{ribeiro2020sketchformer} - replace our whole embedding network with Sketchformer.
    \item w/o CC - remove the coordinate channels of our embedding network.
    \item w/o DF - remove the distance field regression decoder.
\end{itemize}

We train all three networks using our datasets from scratch until convergence. To understand the performance, we first inspect the embedding ability by comparing the reconstructed sketch. 
As shown in Fig.~\ref{pic:ablation}, given the input sketch, Sketchformer can reconstruct a clean but simplified sketch, i.e., the reconstruction either misses a few strokes (e.g., the fingers) or has strokes with inadequate variation (e.g., the wing of the angel). 
Without the coordinate channels (w/o CC) or the distance regression branch (w/o DF), the reconstructed sketch exhibits a significant blur, hard to recognize. 
Instead, our reconstruction almost recovers the input with only a few artifacts when the strokes are cluttered (e.g., the texture of the pineapple or bag), which indicates the higher fidelity of our learned embedding. 

To further examine the segmentation performance with specific embeddings, we link the three embedding variants with our segmentation Transformer (denoted as Sketchformer-Seg, Ours w/o CC, and Ours w/o DF, respectively), and train them on four typical categories (i.e., Airplane, Calculator, Face, and Ice cream) from the SPG dataset. Statistical results are shown in Tab.~\ref{tab:ablation}. Sketchformer-Seg obtained the worst segmentation results due to the under-representation of its embedding. Compared with Ours, the learned embeddings from w/o CC and w/o DF significantly suppress the segmentation ability.

\parag{Parallel Decoding.} Other than the used auto-regressive decoding, the parallel decoding scheme is frequently used in Transformer training and testing. Following~\cite{carion2020end}, we thus change our auto-regressive decoding to the parallel decoding scheme (denoted as Ours-PD) and report the evaluation metrics in Tab.~\ref{tab:ablation}. The learned decoder input embeddings cannot explicitly access the context information, thus leading to notable performance degradation (e.g., 93.6\% vs. 91.6\% of the average CAcc).

\rev{
\parag{Additional Ablation Study.} We have conducted more experiments on a) positional encoding of strokes and b) distance field-only decoding in the embedding learning, c) stroke-based auto-regressive decoding, and d) the group order in auto-regressive decoding, more results statistical and visual results can be found from the supplementary.
}

\section{Discussion}

\begin{table*}[!htb]
\centering
\caption{The statistical comparison before and after the cross-category training on \textbf{SAcc}.}
\vspace{-2mm}
\label{tab:cross_category}
\renewcommand{\arraystretch}{1.3}
\resizebox{1\textwidth }{!}{
\begin{tabular}{c|ccc|c|ccc|c|ccc|c|ccc|c|ccc}
\toprule[0.4mm]
\rowcolor{gray!20}
\multicolumn{4}{c|}{\textbf{Ant}}& \multicolumn{4}{c|}{\textbf{Crab}}& \multicolumn{4}{c|}{\textbf{Pig}}& \multicolumn{4}{c|}{\textbf{Butterfly}}& \multicolumn{4}{c}{\textbf{Airplane}}\\
\hline
Part & before & after&change &Part &before & after&change &Part &before & after&change &Part &before & after&change &Part &before & after&change\\
\hline
Face& 69.3 & 82.4&\textcolor{darkgreen}{+13.1}& Face & 88.4&88.1 &\textcolor{red}{-0.3} &Face &96.0 &96.4 & \textcolor{darkgreen}{+0.4}&Body &98.4 &98.6&\textcolor{darkgreen}{+0.2} &Body &95.2 &95.9&\textcolor{darkgreen}{+0.7}\\
Head & 91.7 &92.8  &\textcolor{darkgreen}{+1.1} & Body & 95.1& 96.2& \textcolor{darkgreen}{+1.1} &Head &97.6 &98.3&\textcolor{darkgreen}{+0.7} &Tentacle &98.6 &99.1&\textcolor{darkgreen}{+0.5} &Front wings &96.3 & 96.7&\textcolor{darkgreen}{+0.4}  \\
Tentacle& 88.7 &91.8&\textcolor{darkgreen}{+3.1} & Legs &96.9 &97.3 &\textcolor{darkgreen}{+0.4}&Body &93.9 &95.0& \textcolor{darkgreen}{+1.1}&Wings & 99.3&99.4& \textcolor{darkgreen}{+0.1}&Windows & 85.0&86.4&\textcolor{darkgreen}{+1.4} \\
Body& 95.9 & 96.4&\textcolor{darkgreen}{+0.5}& Pliers & 96.4&96.1 &  \textcolor{red}{-0.3}&Legs &98.5 &98.7 & \textcolor{darkgreen}{+0.2}& & & & &Tail wings& 87.1& 84.8&\textcolor{red}{-2.3}\\
Legs& 96.2 &96.5&  \textcolor{darkgreen}{+0.3}& & && & Tails& 96.7&98.9 &\textcolor{darkgreen}{+2.2} &&& && & && \\
\hline
Average& 88.4 &92.0  &\textcolor{darkgreen}{+3.6} & Average & 94.2&94.4  &\textcolor{darkgreen}{+0.2}&Average &96.5 &  97.5&\textcolor{darkgreen}{+1.0}& Average &98.8 &99.0 & \textcolor{darkgreen}{+0.2}& Average & 90.9&  91.0& \textcolor{darkgreen}{+0.1}\\
\bottomrule[0.4mm]
\end{tabular}
}

\end{table*}


\begin{figure*}[!htb]
    \centering
    \vspace{-3mm}
    \includegraphics[width=0.95\textwidth]{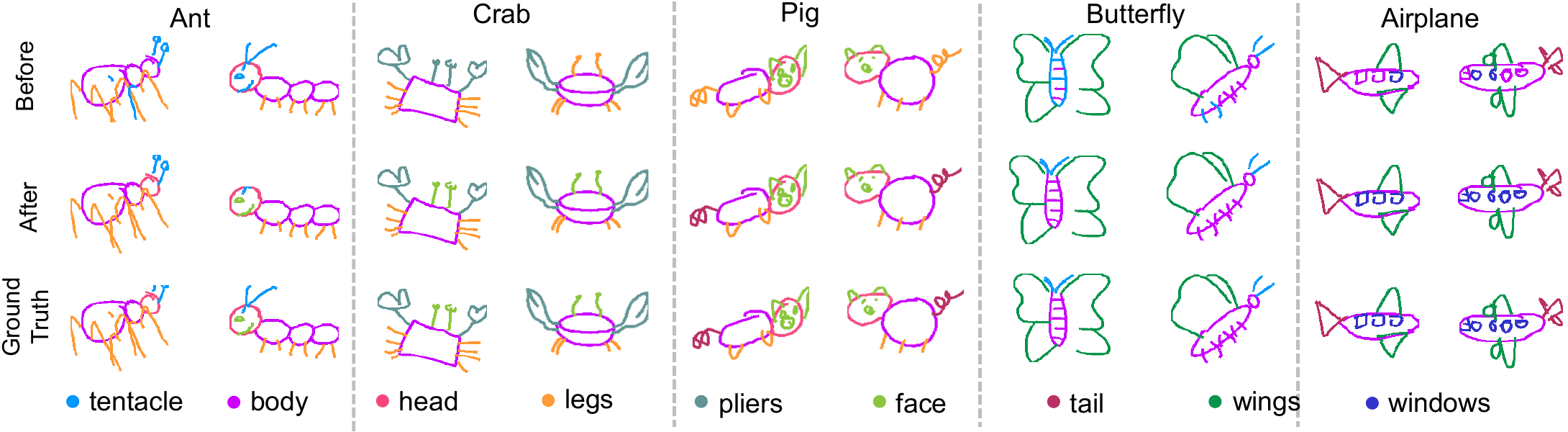}
    \vspace{-4mm}
    \caption{Visual segmentation results before and after cross-category training.}
    \label{pic:cross}
    \vspace{-4mm}
\end{figure*}

\subsection{Cross-Category Semantic Segmentation}
\label{subsec:cross_categ}

For sketch semantic segmentation, cross-category training is challenging due to the diversity of training data categories. However, we hypothesize that if different categories have semantically, geometrically, and positionally similar parts, they should be mutually helpful to each other, and our segmentation Transformer excels at learning such contextual information thus benefiting from cross-category training.
To verify this point, we conducted a preliminary experiment by manually selecting five categories - crab, ant, butterfly, airplane, and pig as the combined training data. After collecting all the parts, we merge the eye, mouth, nose, and ear into a single part -- ''face'' because these parts all semantically and positionally compose the face part, and their stroke frequencies are low. Thus, we use nine parts, i.e., pliers, body, legs, head, tentacle, wings, tail, window, and face in the combined dataset.

After training the network, statistical evaluation is reported in Tab.~\ref{tab:cross_category}, where clear improvements (green color) can be seen from all five categories, especially the ant face category with a remarkable $13.1\%$ improvement, validating the superiority of our method in detecting and exploiting contextual information and the feasibility of cross-category training. Performance degradation is observed from the face and pliers parts of the crab, and the tail wing part of the airplane. Because these parts are only semantically but not positionally or geometrically similar to others, for example, the eyes of some crabs are out of the head, and the tail of an airplane has more complex patterns (e.g., triangles) than a simple curved tail. Qualitative results are presented in Fig.~\ref{pic:cross} with clear visual improvements consistent with statistics.

\subsection{Sementic-aware Data Augmentation}

\begin{table}[!tb]
\centering
\caption{The statistical evaluation of each part before and after using the semantic-aware data augmentation on \textbf{SAcc}.}
\vspace{-2mm}
\renewcommand{\arraystretch}{1.3}
\resizebox{1.0\linewidth}{!}{
\begin{tabular}{c|ccc|c|ccc}
\toprule[0.4mm]
\rowcolor{gray!20}
\multicolumn{4}{c|}{\textbf{Ant}} & \multicolumn{4}{c}{\textbf{Ambulance}}\\
\hline
Part & before & after&change &Part &before & after&change\\
\hline
Eye,mouth& 69.3 & 89.1 &\textcolor{darkgreen}{\textbf{+19.8}} & Bell & 97.5& 97.4& \textcolor{red}{-0.1} \\
Head & 91.7 & 91.9 &\textcolor{darkgreen}{+0.2} & Body & 97.5& 97.9 & \textcolor{darkgreen}{+0.4 } \\
Tentacle& 88.7 & 88.8&\textcolor{darkgreen}{+0.1} & Windows & 58.6& 76.2&\textcolor{darkgreen}{\textbf{+17.6}}\\
Body& 95.9 & 95.8&\textcolor{red}{-0.1}& Cross & 97.1& 96.8 & \textcolor{red}{-0.3} \\
Legs& 96.2 & 96.9&\textcolor{darkgreen}{+0.7}& Wheels & 94.8 & 96.1  &\textcolor{darkgreen}{+1.3 }\\
\hline
Average& 88.4 & 92.5 &\textcolor{darkgreen}{+4.1} & Average & 89.1& 92.9 &\textcolor{darkgreen}{+3.8}\\
\bottomrule[0.4mm]
\end{tabular}
}
\label{tab:data_augmentation}
\end{table}

\begin{figure}[!t]
    \centering
    \vspace{-2mm}
    \includegraphics[width=1.0\linewidth]{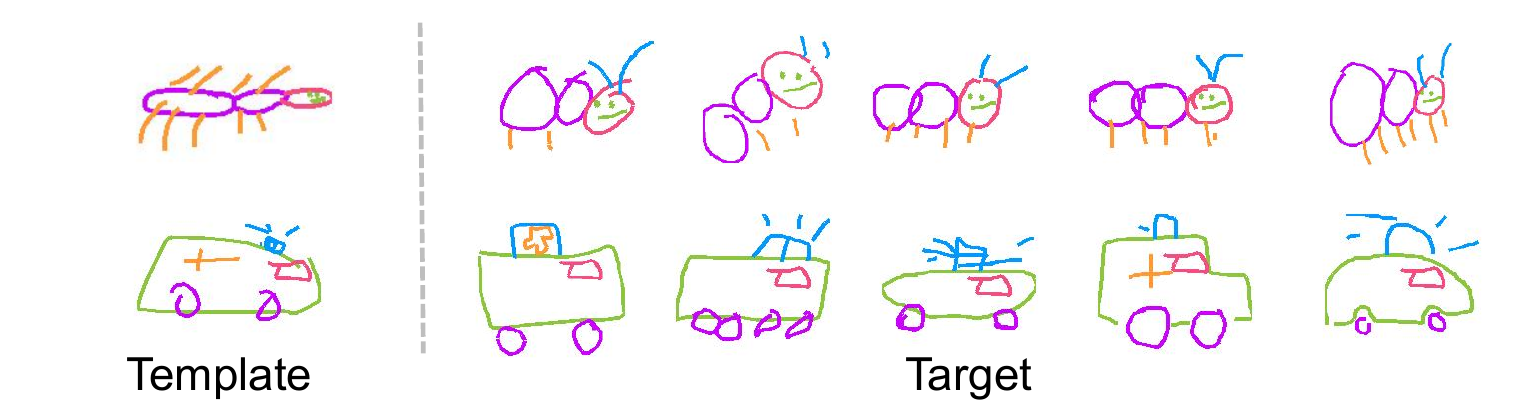}
    \vspace{-7mm}
    \caption{ Given the eye of a template ant or a window of a template ambulance, we copy that part and paste it into the target sketches based on semantic constraints.}
    \label{pic:data_aug_process}
    \vspace{-5mm}
\end{figure}

In the SGP dataset, there is a significant imbalance in terms of the stroke distribution over different parts within certain categories. For example, in the ``ant" category, the strokes for the "eye" part account for only $3.2\%$ of all strokes, and the stroke accuracy for this part is also notably low at $69.3\%$ even with the focal loss. We thus propose a novel semantic-aware data augmentation by copying and pasting. Specifically, as shown in Fig.~\ref{pic:data_aug_process}, we select the ``ant'' and ``ambulance'' categories as the playground since they exhibit the aforementioned imbalance (see Tab.~\ref{tab:data_augmentation}, the before column). We first divide the training examples into two sets with or without the ``eye,mouth'' part, denoted as A and B. We then randomly select a template example from A and copy the interested part to B based on the semantic constraints, e.g., the ``eye,mouth'' part must appear inside the head. Scaling is necessary to fit the added part to target examples in set B. Additionally, we introduce part-level rotation and offset perturbations to further improve the diversity. By copying and pasting the rare part, we improve its occurrence to around $50\%$ in the whole category. For the ambulance, a similar process is applied to the ``window'' part.

\begin{figure}[!t]
	\centering
    \includegraphics[width=1.0\linewidth]{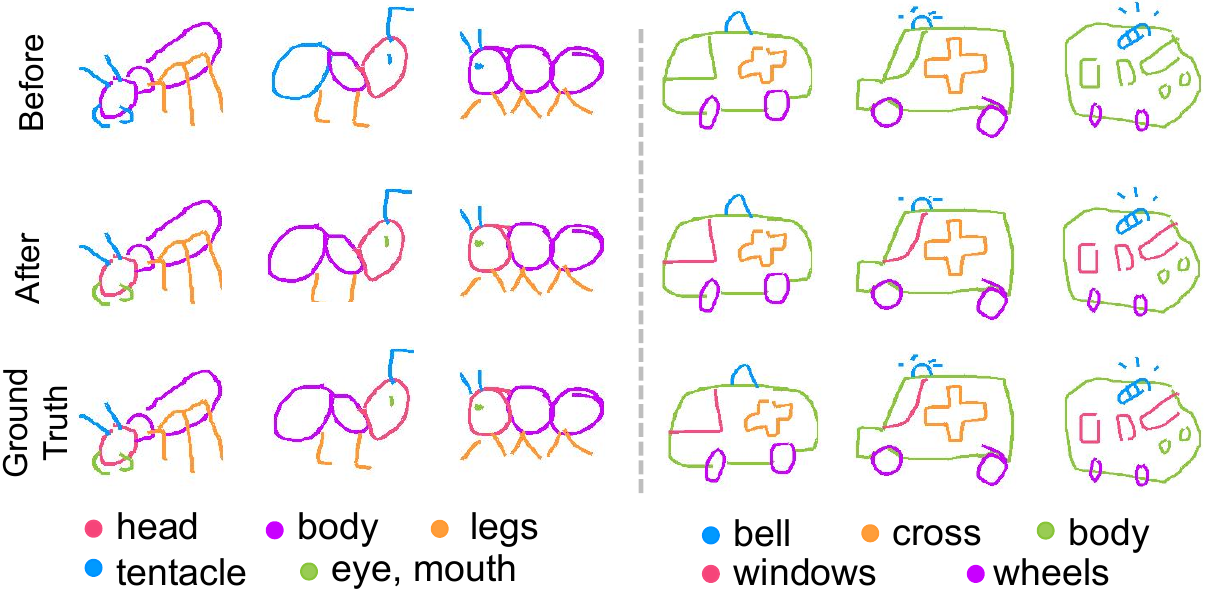}
    \vspace{-6mm}
	\caption{Visual results of the semantic segmentation with the novel data augmentation.}
    \vspace{-4mm}
	\label{pic:data_aug2}
\end{figure}

We use the augmented data to train our segmentation network. Statistical results are presented in Tab.~\ref{tab:data_augmentation}, where remarkable improvements of the ``eye,mouth'' and ``window'' parts are observed, indicating the efficiency of the semantic-aware data augmentation. Visual results shown in Fig.~\ref{pic:data_aug2} support the statistical observation.

\subsection{Invariance Test}
Following~\cite{yang2021sketchgnn,zheng2023sketch}, we also conducted invariance tests in terms of anti-rotation and anti-offset. As shown in Tab.~\ref{tab:inv_test} in the supplementary, our performance drops when increasing the rotation angle and the offset distance of the testing examples, but we still obtain superior accuracy than SketchGNN~\cite{yang2021sketchgnn} and Sketch-Segformer~\cite{zheng2023sketch}, which indicates the excellent robustness of our approach. See the supplementary for detailed configuration and results.

\section{Conclusion}
In this paper, we proposed \methodName, for sketch semantic segmentation,
achieving the best segmentation accuracy over SOTA methods. 
Comprehensive evaluations validate our superior performance, and preliminary experiments on cross-category training and semantic-aware data augmentation suggest inspiring research directions. 
However, our embedding network is not perfect and cannot encode strokes with rapid and large variations, while our segmentation Transformer fails to label strokes that are highly cluttered, e.g., the hairs and eyelashes. 

{
    \small
    \bibliographystyle{ieeenat_fullname}
    \bibliography{main}
}

\clearpage
\setcounter{page}{1}
\maketitlesupplementary

\section{More Results}
\begin{figure*}[!t]
    \centering
    \includegraphics[width=1\textwidth]{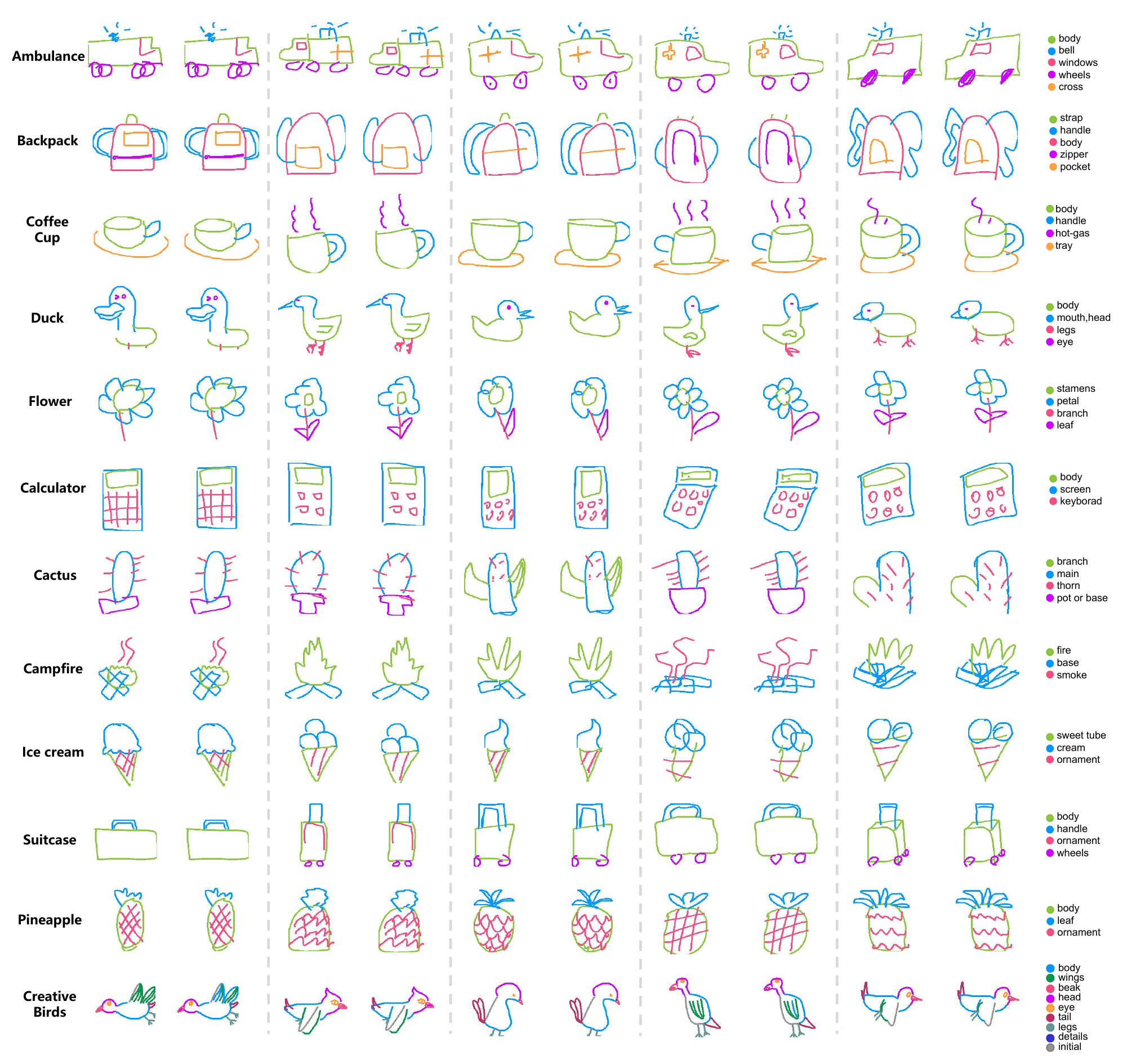}
    \vspace{-8mm}
    \caption{More visual results on benchmarking datasets. For each category, we select five sketch instances, and for each sketch, the ground truth (left) and our segmentation results (right) are displayed. 
    }
    \label{suppl_pic:more_result}
\end{figure*}

Figure \ref{suppl_pic:more_result} presents more segmentation results of our \methodName from the SPG and CreatureSketch (CS) datasets. Our method is robust to sketches with various levels of detail.

\parag{Statistical Analysis.}
Table \ref{tab:main_stats} in the main paper demonstrates the comparison with three competitive methods, and our \methodName outperforms these competitors on all three metrics (except for the face category).
Specifically, \methodName, in particular, exhibits an average improvement of 1.1\% in Stroke Accuracy (SAcc), 0.4\% in Group Accuracy (GAcc), and 1.4\% in Component Accuracy (CAcc) over Sketch-Segformer, which stands out as the most effective among alternative methods. This discrepancy can be attributed to Sketch-Segformer's reliance on absolute coordinates encoded within its graph representation, which, unfortunately, struggles to encapsulate essential structural information.
Furthermore, the proposed \methodName demonstrates an average improvement of 1.6\% in SAcc, 0.6\% in GAcc, and 1.8\% in CAcc compared to SketchGNN, which similarly relies on absolute coordinates to represent graph-based sketches. However, like Sketch-Segformer, SketchGNN also struggles to accurately capture the structural information inherent in strokes.
Ultimately, the proposed \methodName showcases significant superiority over SPGSeg, a sequence-based method, with an average outperformance of 17.4\% in SAcc, 10\% in GAcc, and 18.7\% in CAcc. SPGSeg employs sequential encoding of sketches using relative coordinates and stroke point pen states. However, it overlooks the proximity of strokes, contributing to its comparative shortcomings.

\subsection{Additional Ablation Study}
\begin{figure}[!htb]
    \centering
    \includegraphics[width=0.9\linewidth]{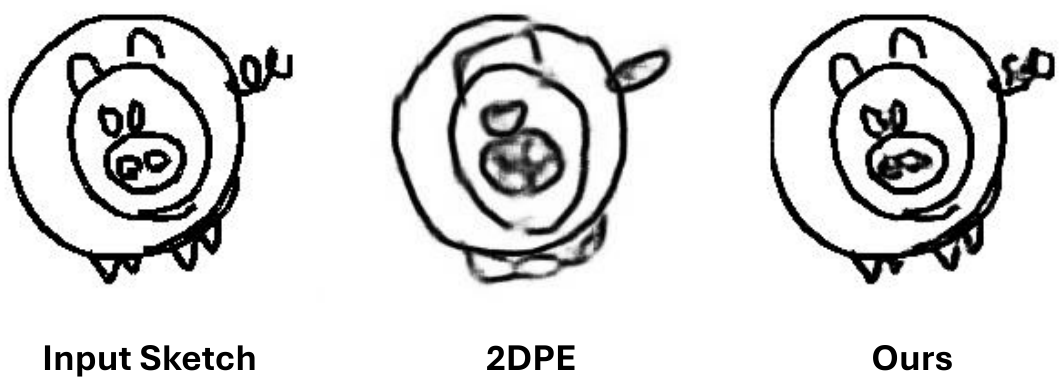}
    \vspace{-3mm}
    \caption{Sketch reconstruction from 2DPE and Ours.}
    \label{fig:embed_2DPE}
\end{figure}

\begin{figure}[!htb]
    \centering
    \includegraphics[width=0.9\linewidth]{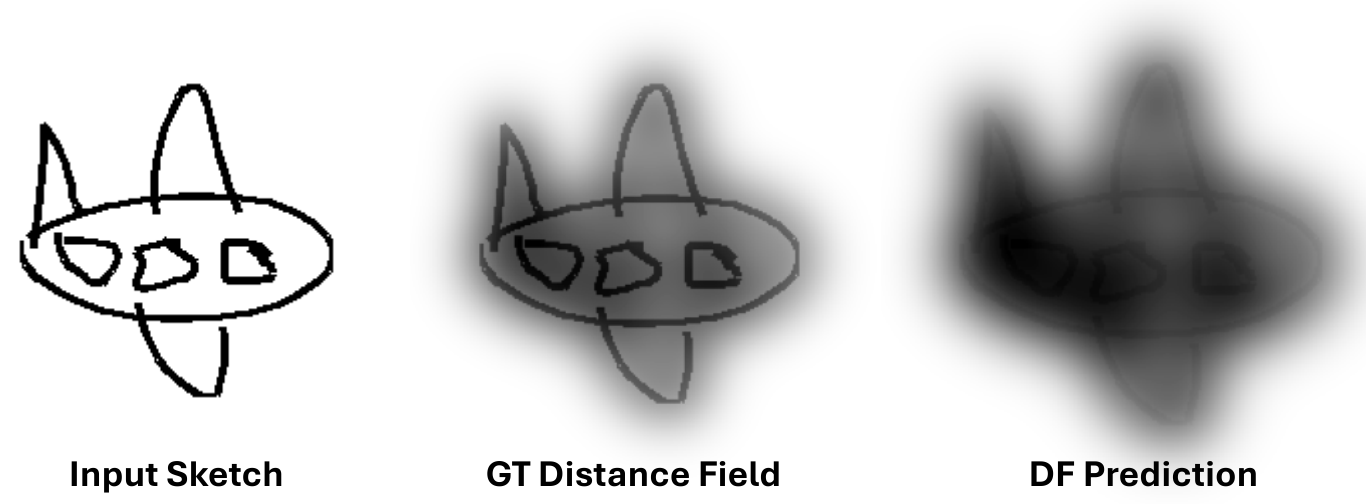}
    \vspace{-3mm}
    \caption{Distance field prediction from DF.}
    \label{fig:df_pred}
\end{figure}

\rev{
\parag{Positional Encoding.} In stroke embedding learning (\cref{subsec:stroke_embedding}), we have used two additional coordinate channels to augment the stroke information. To understand better its effectiveness, we replaced the 2D coordinates with the popular 2D sinusoidal positional encoding (denoted as 2DPE and Ours-PE for embedding and segmentation).
The stroke reconstruction result and the evaluation statistics are shown in Fig. \ref{fig:embed_2DPE} and Tab. \ref{tab:addi_abl}, respectively, where the reconstructed sketch is blurry, and the segmentation metrics are all significantly inferior to ours.

\parag{Distance Field-only Embedding.} To further validate the efficacy of the distance field prediction branch, we have trained the embedding network with only the distance field (denoted as DF). The predicted distance field is shown in Fig. \ref{fig:df_pred}, where individual strokes are barely recognized. Besides, we have linked the embedding network with our segmentor (denoted as Ours-DF), and report the evaluation metrics in Tab. \ref{tab:addi_abl}. The results are inferior to Ours and even worse than Ours w/o DF because the sketch instead of a dense and rough approximation is the key to the segmentation task.

\parag{Stroke-based Decoding.} The design philosophy of the group-based prediction and its effectiveness are discussed in \cref{subsec:seg_trans} and \cref{subsec:abl_study}. We further experiment with stroke-based auto-regressive decoding (denoted as Ours-S) since it is more intuitive. Statistics are shown in the Tab. \ref{tab:addi_abl}, where our method 
achieved a remarkable $4.4\%$ increase in terms of average SAcc. 
Besides, group-based prediction is more efficient, e.g., the average inference time on the airplane category is two times faster (0.73s vs. 1.86s).
 
\parag{Group Order in Auto-regressive Decoding.} By design, the stroke order serving as the positional encoding in the Transformer encoder does not matter the decoder prediction, however, the group order matters.
By default, we empirically use the more intuitive stroke frequency-based descend order.
We have tried two alternatives: stroke frequency ascend order (denoted as Ours-InvG), and a random order (denoted as Ours-RanG). The statistics are shown in Tab. \ref{tab:addi_abl},  where ours archives the best while the other two are on par with ours.
}

\begin{table*}[t]
\centering
\caption{Statistical results of additional ablation studies on the SPG dataset.}
\vspace{-2mm}
\label{tab:addi_abl}
\renewcommand{\arraystretch}{1.2}
\resizebox{1.\linewidth }{!}{
\begin{tabular}{c|ccc|ccc|ccc|ccc|ccc}
\toprule[0.4mm]
\multirow{2}{*}{Method} & \multicolumn{3}{c|}{Airplane}& \multicolumn{3}{c|}{Calculator}& \multicolumn{3}{c|}{Face }& \multicolumn{3}{c|}{Ice cream}& \multicolumn{3}{c}{Average} \\
\cline{2-16}

& SAcc& GAcc & CAcc& SAcc& GAcc & CAcc& SAcc& GAcc & CAcc& SAcc& GAcc & CAcc & SAcc& GAcc & CAcc\\

\hline

\rowcolor{lightgray}
Ours-PE  & 89.3  & 91.9 & 85.6 & 96.1& 95.7& 96.8&93.4 & 95.4 &93.4 & 93.1 & 94.5 &91.2 & 93.0 & 94.4 & 91.8 \\

Ours-DF &  83.4 & 86.6 & 81.3 & 90.1 & 91.7 & 87.2& 89.4 & 89.2& 86.3 & 86.7 & 89.3 & 84.2 & 87.4 & 89.2 & 84.8\\


\rowcolor{lightgray}
Ours-S & 88.6 & 91,1 & 84.7 & 94.4 & 95.8 & 93.6 &92.1 & 94.6 & 91.7 & 91.8 &  92.3& 89.6& 91.7  &93.5  &  89.9\\

Ours-InvG & 92.6 &94.0  &88.6  &98.6 & 98.1&97.0 &95.9 &97.9 & 93.8& 95.2& 96.1 &91.7 & 95,6 & 96.5 & 92.8 \\

\rowcolor{lightgray}
Ours-RanG &92.8 & 94.1 &88.7 &98.7 & 98.2&97.1 & 96.1& 98.0& 94.1& 95.4  &  96.2& 91.9& 95.8 &96.8  &93.0  \\
\hline

Ours & \textbf{93.2} &\textbf{94.9} & \textbf{89.5}& \textbf{99.2} &\textbf{98.7} & \textbf{97.5}& \textbf{96.4}& \textbf{98.4} & \textbf{94.8} &\textbf{95.9}& \textbf{96.5} & \textbf{92.4} &\textbf{96.1} & \textbf{97.1}
& \textbf{93.6} \\
\bottomrule[0.4mm]
\end{tabular}
}
\end{table*}

\subsection{Invariance Test}
Following~\cite{yang2021sketchgnn,zheng2023sketch}, we also conducted invariance tests in terms of the anti-rotation and anti-offset ability of our approach. The experiments are conducted on four typical categories - Airplane, Calculator, Face, and Icecream, and statistical results are reported in Tab.~\ref{tab:inv_test}.

\parag{Anti-rotation test.} Regarding the anti-rotation test, we adopted an identical experimental setup to previous studies. This setup involved the inclusion of seven distinct rotation angles (i.e., $-45^{\circ}$, $-30^{\circ}$, $-15^{\circ}$, $0^{\circ}$, $+15^{\circ}$, $+30^{\circ}$, and $+45^{\circ}$) added to the entire sketch.
The findings presented in Table \ref{tab:inv_test} reveal a trend wherein the performance of both competitive methods declines with increasing rotation angles. Notably, the proposed \methodName exhibits a superior mean performance and a narrower standard deviation when compared to other models. This outcome underscores the model's exceptional reliability, particularly in handling sketches subjected to significant degrees of rotation. 

\parag{Anti-offset test.} For the offset test, given a testing sketch, we first calculate the diagonal length $d$ of the bounding box, and for each stroke, we randomly sample an offset from a uniform distribution - $\left( \Delta x,\Delta y \right) \sim \mathcal N(0,\sigma^2)$, where $\sigma$ is set at $0.05d$, $0.1d$, $0.15d$, and $0.20d$, respectively.
As expected, our performance drops when increasing the offset distance, but we still obtain superior accuracy than SketchGNN~\cite{yang2021sketchgnn} and Sketch-Segformer~\cite{zheng2023sketch} at each variation, which indicates the excellent robustness of our approach.

The results from both tests strongly suggest that ContextSeg possesses greater robustness, showcasing its ability to sustain segmentation accuracy despite offsets or rotations. This signifies its strength in effectively managing spatial variations within the data.

\begin{table*}[!htb]
\centering
\caption{Statistical comparison of the invariance tests on four representative categories.}
\vspace{-2mm}
\renewcommand{\arraystretch}{1.3}
\resizebox{1\textwidth }{!}{
\begin{tabular}{c|c|cccc|cccc|cccc}
\toprule[0.4mm]

&\multirow{2}*{Angle/Distance} & \multicolumn{4}{c|}{\textbf{SketchGNN}~\cite{yang2021sketchgnn} }  & \multicolumn{4}{c|}{\textbf{Sketch-Segformer}~\cite{zheng2023sketch} }  & \multicolumn{4}{c}{Ours} \\
\cline{3-14}
&& Airplane & Calculator &  Face & Ice cream& Airplane & Calculator &  Face & Ice cream& Airplane & Calculator &  Face & Ice cream \\
\hline
\hline

\multirow{9}*{ \rotatebox{90}{\textbf{Rotation Test}}}&$-45^{\circ}$ &87.7 &92.4 &90.7 &88.4 & 88.2 &93.1 &90.2 &87.5 & \textbf{90.1} & \textbf{94.9}&\textbf{91.7} &\textbf{90.8} \\
&$-30^{\circ}$ &89.4 &94.7 &93.2 &91.3 &89.9 &95.4 &92.1 &90.5 & \textbf{91.0}  & \textbf{97.2} & \textbf{94.1} & \textbf{92.1}\\

&$-15^{\circ}$ &90.4& 97.6& 95.7& 93.4& 91.3&98.1 &94.8 & 92.1 & \textbf{92.6 } & \textbf{98.5} & \textbf{95.9} & \textbf{94.3}\\
&0 & 91.1 & 98.4 & 96.2 & 94.5 & 92.4 & 98.5 & 97.2 & 94.4 & \textbf{93.2}  & \textbf{99.2} & \textbf{96.4} & \textbf{95.9}\\

&$+15^{\circ}$ &90.1 & 97.1 &95.2 &92.9 &91.1 &98.3 &94.1 &92.7 &  \textbf{92.9}  & \textbf{98.2} & \textbf{95.5} & \textbf{94.1}\\
&$+30^{\circ}$ &89.2 &94.3 &93.1 &90.8 &90.0 &95.9 &92.7 &90.1 & \textbf{91.3 } & \textbf{97.4} & \textbf{93.9} & \textbf{92.3}\\

&$+45^{\circ}$ &87.2 &92.1 &90.1  &87.8 &88.5 &93.7 &89.6 & 87.1&  \textbf{89.5} & \textbf{95.3}&\textbf{91.3 }&\textbf{90.4} \\
\cline{2-14}
&Average&  89.3 & 95.2 & 93.5 & 91.3&90.2 &  96.1 & 93.0 & 90.6 & \textbf{91.5}& \textbf{97.2} & \textbf{94.1}& \textbf{92.8} \\

&Standard Deviation& \textbf{1.4}& 2.5 & 2.4 & 2.5& 1.5&  2.2& 2.7 & 2.7& \textbf{1.4}  &\textbf{1.6}  &\textbf{2.0} &\textbf{2.0}\\

\hline
\hline

\multirow{7}*{ \rotatebox{90}{\textbf{Offset Test}}}&0 & 91.1 & 98.4 & 96.2 & 94.5 & 92.4 & 98.5 & 97.2 & 94.4 & \textbf{93.2}  & \textbf{99.2} & \textbf{96.4} & \textbf{95.9}\\

&$0.05\sigma$ &90.3 &96.5 &92.4 &91.4 &90.5 &96.7 &92.7 &91.3 & \textbf{91.1} & \textbf{97.4} & \textbf{93.5} & \textbf{92.1}\\

&$0.1\sigma$ &88.1 &93.0 &90.4 &89.2 &88.7 &93.4 &91.0 & 88.4&  \textbf{89.1}& \textbf{94.1}& \textbf{91.9} & \textbf{90.1}\\

&$0.15\sigma$ &85.5 &89.4 &87.1 &86.1 &85.2 &88.7 &87.6 & 85.3& \textbf{86.4} & \textbf{91.4} & \textbf{88.4} & \textbf{87.8} \\

&$0.20\sigma$ &79.9 &85.4 &81.0 &82.5 &78.6 &84.2 &82.1 &81.9&  \textbf{82.4}& \textbf{87.5} &\textbf{83.7} & \textbf{84.5} \\
\cline{2-14}

&Average&87.0 &  92.5&89.4& 88.7&87.1&92.3&90.1& 88.3&  \textbf{88.4}& \textbf{93.9 }&\textbf{90.8 }& \textbf{90.1}\\

&Standard Deviation&4.5&5.3& 5.7& 4.6 &5.4 &5.9&5.7& 4.9& \textbf{4.2} &  \textbf{4.7} & \textbf{4.9}&\textbf{4.3}\\
\bottomrule[0.4mm]
\end{tabular}
\label{tab:inv_test}
}
\end{table*}


\begin{figure}[!htb]
    \centering
    \includegraphics[width=1.0\linewidth]{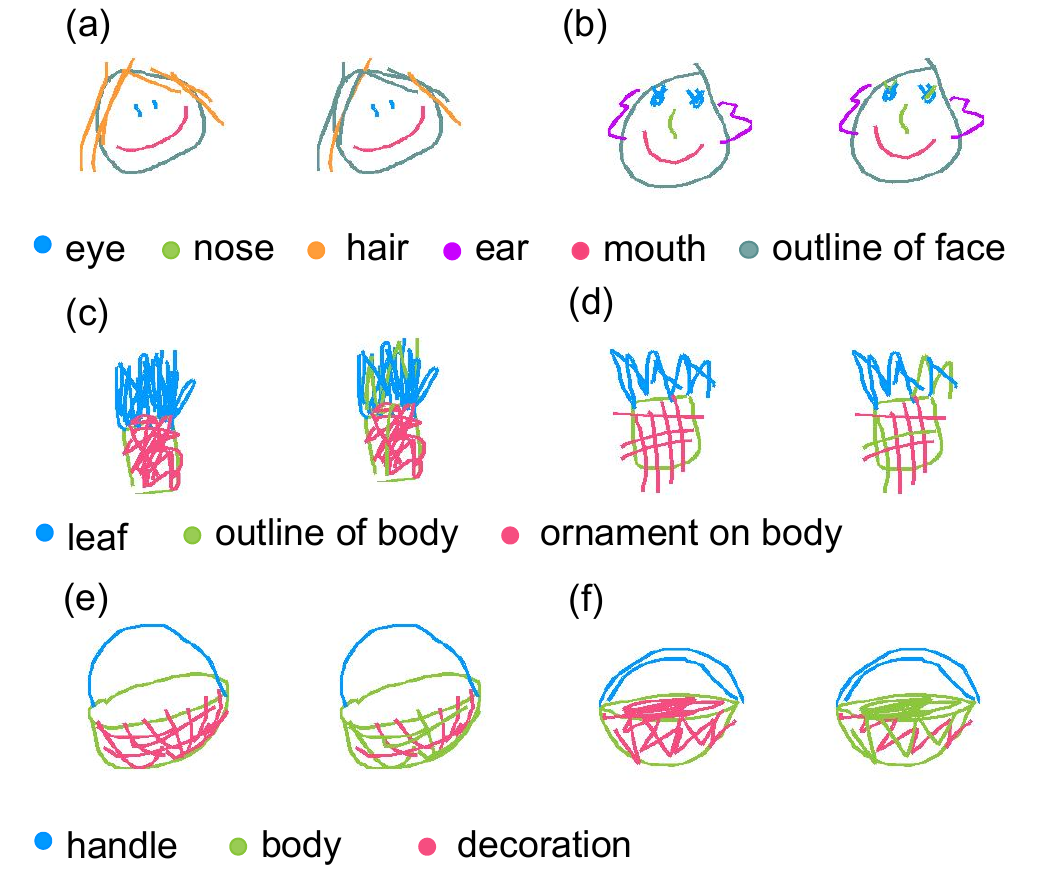}
    \vspace{-8mm}
    \caption{Exemplar results with imperfect segmentation. For each case, the ground truth (left) and our prediction (right) are shown.}
    \label{suppl_pic:failure_cases}
\end{figure}

\subsection{Failure Cases}
Figure \ref{suppl_pic:failure_cases} demonstrates several failure cases of our method. 
The imperfection primarily arises from two contributing factors.
Firstly, our segmentation Transformer encounters difficulties in accurately labeling strokes within heavily overlapped areas. For instance, in Fig.~\ref{suppl_pic:failure_cases} (a) and (b), distinguishing between hair and eyelash strokes becomes challenging due to their dense concentration, leading to segmentation inaccuracies. Similarly, in Fig.~\ref{suppl_pic:failure_cases} (e) and (f), the intricate decorations on the basket pose challenges as our segmentation model erroneously categorizes them as part of the body.
Secondly, our embedding network encounters limitations in encoding strokes characterized by rapid and substantial variations. For instance, in Fig.~\ref{suppl_pic:failure_cases} (c) and (d), the leaves and ornamentation of the pineapple exhibit strokes with rapid fluctuations. Such variations can result in suboptimal embeddings, potentially causing misinterpretations for our segmentation Transformer. Consequently, certain strokes might be inaccurately classified as part of the body.

\section{Implementation Detail}
\label{subsec:imp_detail}

\begin{figure*}[!tb]
\centering

\begin{minipage}[t]{0.49\textwidth}
\centering
\includegraphics[width=\linewidth]{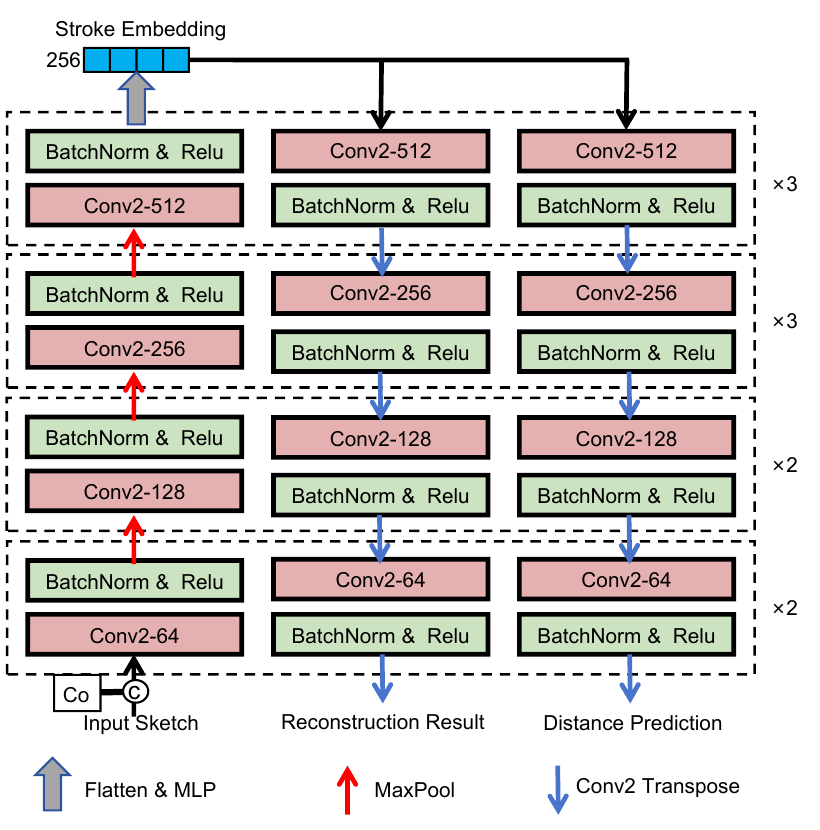}
{\small (a) Embedding Network.}
\end{minipage}
\hfill
\begin{minipage}[t]{0.49\textwidth}
\centering
\includegraphics[width=\linewidth]{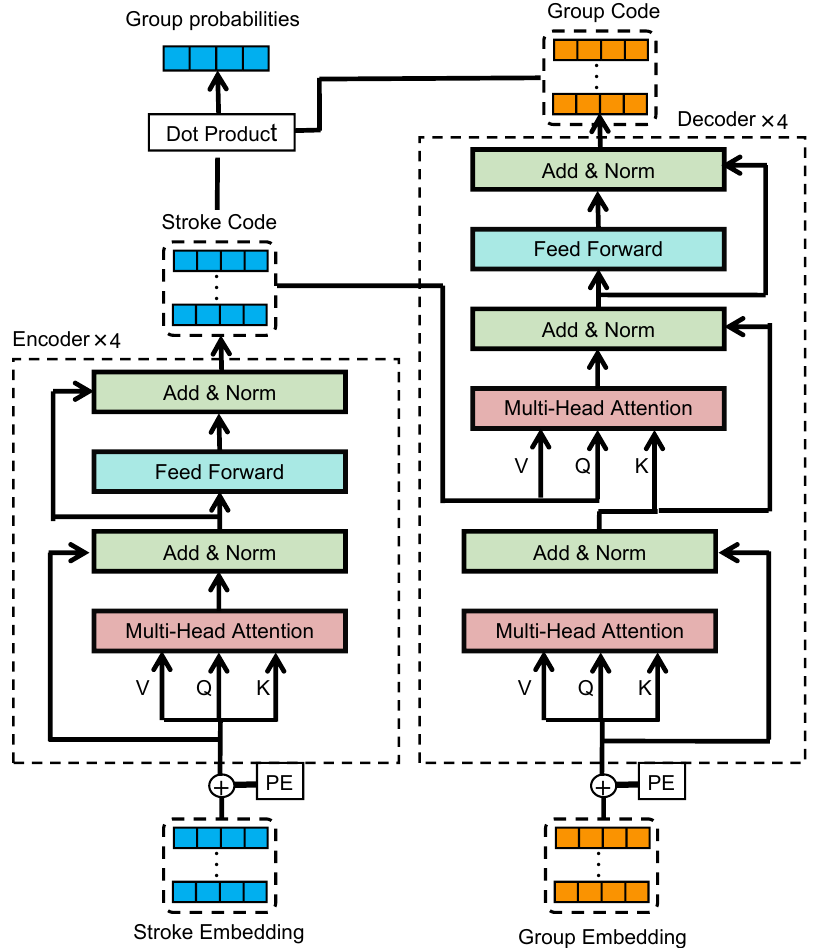}
{\small (b) Segmentation Transformer.}
\end{minipage}
\vspace{-2mm}
\caption{Detailed structure of our embedding and segmentation networks.}
\label{suppl_pic:networks}
\end{figure*}

All experiments were conducted on a single Nvidia RTX3090 GPU. We implement our networks using Tensorflow, and detailed network structures are shown in Fig.~\ref{suppl_pic:networks}.

The embedding network has an encoder-decoder structure, accepting the grayscale sketch input augmented with x and y coordinate channels.
Specifically, the encoder comprises 10 layers grouped into four segments, each characterized by distinct feature dimensions (i.e., 64, 128, 256, and 512), resulting in a stroke embedding of $256d$. Both decoder branches share an identical encoder structure, working symmetrically to transform the stroke embedding into sketch reconstruction and the distance map.

The segmentation Transformer has 4 attention layers in both the encoder and decoder, each layer has 4 attention heads and the dropout rate is 0.4. 

\parag{Network Training.} We first train the embedding network until convergence, which takes around 15 hours with a batch size of 64. Then, the segmentation Transformer is trained until convergence taking around 10 hours with a batch size of 8. Adam~\cite{kingma2014adam} optimizer was used in both network training with a fixed learning rate $10^{-4}$ and other default parameters. 

\parag{Teacher Forcing Gap.} Teacher-forcing is widely used for Transformer training. However, it introduces the exposure bias issue by feeding the ground truth context to the decoder at training time while exploiting the inferior prediction at testing time. To overcome the teacher-forcing gap, in our case, we follow \cite{mihaylova2019scheduled} to forward the decoder twice to mix the predicted group of strokes with the ground truth group of strokes. The ratio of the ground truth strokes gradually decreases from $100\%$ to $20\%$ along with the training process.

\section{Dataset Details}
\parag{Evaluation Metrics.}
The three metrics we used in the paper are defined as:
\begin{itemize}
    \item \textbf{Stroke Accuracy (SAcc)} calculates the percentage of correctly labeled strokes. For a point-based stroke representation, if a minimum of $75\%$ of its points are correctly labeled, then the stroke is correctly labeled.
    \item \textbf{Grouping Accuracy (GAcc)} measures the accuracy of the group-based classification task in our segmentation Transformer. Suppose the ground truth classification labels are stored in a binary matrix $M^{\mathbbm{S}\times \mathbbm{C}}$, where $\mathbbm{S}$ is the total number of strokes and $\mathbbm{C}$ is the total number of groups/categories. $M_{i,j}=1$ if and only if stroke $s_i$ belongs to $g_j$. The Transformer predicts $M'$ given a sketch $s_i$, we thus calculate the grouping accuracy as: 
    \begin{equation}
        GAcc = \frac{1}{\mathbbm{S}\times \mathbbm{C}} \left|{M}-{M'}\right|.
    \end{equation}
    \item \textbf{Component Accuracy (CAcc)} measures the percentage of correctly labeled categories. A category is deemed accurately labeled if a minimum of $75\%$ of its strokes receive the correct labels.
\end{itemize}

\parag{Data Augmentation.} 
To enrich the diversity of the dataset and improve the robustness of the trained networks, we apply both stroke-level and sketch-level data augmentations. For the former, we rotate, scale, and add a positional perturbation to one or more strokes within a sketch. While, for the latter, we rotate and scale the sketch image, and randomly discard strokes from sketch images as well.

\end{document}